\documentclass{article}
\usepackage[preprint]{neurips_2026}
\usepackage[T1]{fontenc}
\usepackage{microtype}
\usepackage{amsmath}
\usepackage{booktabs,tabularx}
\usepackage{graphicx}
\usepackage{xcolor}
\usepackage{xurl}
\usepackage{hyperref}

\definecolor{linkblue}{HTML}{204E69}
\definecolor{commitment}{HTML}{8B311D}
\hypersetup{
  colorlinks=true, linkcolor=linkblue, citecolor=linkblue, urlcolor=linkblue,
  pdftitle={Same Change, Different Meaning: A Paired Audit of Epistemic Drift in LLM Rewrites},
  pdfauthor={Aliaksei Kaliutau}
}
\setcitestyle{authoryear,round}
\newcommand{\changed}[1]{\textcolor{commitment}{\textbf{#1}}}
\newcommand{\bare}{\textsc{Bare}}
\newcommand{\rewrite}{\textsc{Rewrite}}
\newcommand{\checklist}{\textsc{Checklist}}
\newcommand{\quoted}[1]{\textquotedblleft#1\textquotedblright}
\newcommand{\NValidRewrites}{80}
\newcommand{\NOutsideLength}{16}
\newcommand{\MedianRatio}{99.1}
\newcommand{\MedianSequenceSimilarity}{0.19}
\newcommand{\NCompleteJudgments}{924}
\newcommand{\NHeadingApprovals}{909}

\newcommand{\NUnavailable}{7}
\newcommand{\NInterrupted}{5}
\newcommand{\NInvalidSchema}{24}
\newcommand{\NContextReviewed}{24}
\newcommand{\NClearContext}{22}
\newcommand{\NAdditionalChanges}{20}
\newcommand{\ResultMarginalRows}{%
  Gemma 4 31B & 34/76 & 61/77 & 19/80 \\
  GLM 5.3 Flash & 0/74 & 18/78 & 0/77 \\
  DeepSeek V4 Flash & 12/73 & 60/77 & 1/78 \\
  Qwen 3.8 27B & 1/78 & 27/78 & 0/78 \\
}
\newcommand{\ResultControlRows}{%
  Gemma 4 31B & 76 / 0 / 76 & 77 / 0 / 77 & 80 / 0 / 80 \\
  GLM 5.3 Flash & 71 / 0 / 74 & 74 / 0 / 78 & 69 / 0 / 77 \\
  DeepSeek V4 Flash & 73 / 0 / 73 & 77 / 0 / 77 & 78 / 0 / 78 \\
  Qwen 3.8 27B & 78 / 0 / 78 & 78 / 0 / 78 & 78 / 0 / 78 \\
}
\newcommand{\ResultPairedRows}{%
  Gemma 4 31B & \bare & 74$^{\dagger}$ & 26/0 & +35.1 [24.3, 45.9] & $ 1.49 \times 10^{-7} $ & [28.8, 37.5] \\
  Gemma 4 31B & \checklist & 77 & 42/0 & +54.5 [42.9, 64.9] & $ 3.18 \times 10^{-12} $ & [52.5, 56.3] \\
  GLM 5.3 Flash & \bare & 73$^{\dagger}$ & 17/0 & +23.3 [13.7, 32.9] & $ 1.53 \times 10^{-5} $ & [15.0, 25.0] \\
  GLM 5.3 Flash & \checklist & 75$^{\dagger}$ & 18/0 & +24.0 [14.7, 34.7] & $ 1.53 \times 10^{-5} $ & [18.8, 25.0] \\
  DeepSeek V4 Flash & \bare & 70$^{\dagger}$ & 42/0 & +60.0 [48.6, 71.4] & $ 3.18 \times 10^{-12} $ & [51.3, 63.8] \\
  DeepSeek V4 Flash & \checklist & 75$^{\dagger}$ & 57/0 & +76.0 [66.7, 85.3] & $ 1.11 \times 10^{-16} $ & [71.3, 77.5] \\
  Qwen 3.8 27B & \bare & 76 & 26/0 & +34.2 [23.7, 44.7] & $ 1.49 \times 10^{-7} $ & [30.0, 35.0] \\
  Qwen 3.8 27B & \checklist & 76 & 26/0 & +34.2 [23.7, 44.7] & $ 1.49 \times 10^{-7} $ & [31.3, 36.3] \\
}
\newcommand{\ResultMentionRows}{%
  Gemma 4 31B & 0 / 61 \\
  GLM 5.3 Flash & 5 / 18 \\
  DeepSeek V4 Flash & 8 / 60 \\
  Qwen 3.8 27B & 13 / 27 \\
}

\title{Same Change, Different Meaning:\\A Paired Audit of Epistemic Drift in LLM Rewrites}
\author{Aliaksei Kaliutau\\
  Monodromy Labs\\
  \texttt{aliaksei.kaliutau24@alumni.imperial.ac.uk}}

\begin{document}
\raggedbottom
\maketitle

\begin{abstract}
Language models are widely used to polish grammar and style in scientific writing,
but rewriting can introduce subtle semantic errors.
We introduce Semantic Audit, a paired test of meaning-preservation checks, to answer
the question: does a judge model give the same verdict on a single semantic edit when the surrounding prose is rewritten?  On 80 source topics held out from an earlier study, we
plant one change in modality, frequency, or precision and retain its full sentence
verbatim in a broader rewrite. Four model endpoints judge the bare edit, the
broader rewrite, and the same rewrite with a short commitment checklist. 
Our audit exposes a gap between preserving a passage's overall message 
and preserving its commitments.
Across four judges, rewriting increased false approval by 23.3--60.0
percentage points on complete pairs. The checklist reduced it by 24.0--76.0
points. The eight complete-pair tests all had Holm-adjusted \(p<.001\),
and the direction of each effect survived every assignment of the 36
missing outcomes. Three contrasts met the prespecified coverage and
control criteria.

\end{abstract}

\section{Introduction}

Over the past three years, language models have seen growing use in the final
stage of research: writing papers. Many scientists use them to draft text and
summaries from experimental data, while others rely on them for final stylistic
polishing. Compared with other uses, research writing sets a much higher bar
for preserving meaning and accuracy.

Consider a passage reporting "\emph{roughly} 1.5 percent water" by weight. An AI
editor changes it to "\emph{exactly} 1.5 percent". A true value of 1.49 percent
could fit the first claim while falsifying the second. The edit adds precision
without adding a new measurement, which is easy to spot side by side. But what
happens when an AI editor caught itself of this drift, but instead of rolling-back this
change decided to rewrite the rest of the passage as well to match the changed meaning?

Surrounding text can pull in two directions: emphasizing a nested claim or
weakening and diluting it. Readers therefore inherit not just the topic, but
the boundaries of the claim. In pragmatics and discourse analysis, asserting a
claim narrows the possibilities compatible with what participants take for
granted \citep{stalnaker1978}. A reader of \quoted{the effect may persist} can
still allow that it disappears, whereas a reader of \quoted{the effect will
persist} is asked to set that possibility aside. Strengthening a claim adds an
evidential burden, even if the stronger claim happens to be true
\citep{grice1975}. Quantifiers behave similarly: \emph{often} allows exceptions
that \emph{always} rules out. To test fidelity, we can look for a situation
that the source permits and the rewrite rules out---without needing to know
which situation actually occurred. These subtle distinctions can be easy to
miss for models trained largely on fiction, public posts, and general prose.

This burden is easy to lose track of when people rely on each other's expertise.
\citet{hardwig1985} argues that such dependence is ordinary in science because
even experts rely on evidence they cannot all inspect themselves. Editing adds
one more step between evidence and reader. In one sample of coverage of
correlational health research, 81\% of news stories overstated causal claims
when the university release overstated them, compared with 18\% when the
release avoided that overstatement \citep{sumner2014}.
When an automated check certifies a rewrite, the next reader
or system may see the certification and never see the source qualifier. Source
fidelity is still testable at that handoff, even when a full audit of the
underlying evidence is impractical. The stakes are similar in model evaluation.
If an experiment claims to vary style alone, accepting a changed commitment
weakens the interpretation of the experiment's results.

We audit the checker by giving it two versions of the same planted edit
(Figure~\ref{fig:design}). One
changes a short span in the original passage. The other rewords the surrounding
prose while keeping the edited sentence verbatim. Both should be rejected under
the stated meaning-preservation criterion. To test a lightweight intervention,
we add a checklist of commitments to the latter prompt and keep the candidate
texts and their order fixed. The design yields a paired estimate of the change
in false approval and a one-component ablation of the checklist.

\begin{figure}[t]
  \centering
  \setlength{\tabcolsep}{6pt}
  \small
  \begin{tabularx}{\linewidth}{@{}p{.19\linewidth}X@{}}
    \toprule
    Source passage & \ldots a ringwoodite inclusion holding \textbf{roughly} 1.5 percent water by weight \ldots \\
    \midrule
    \bare & Change \textbf{roughly} to \changed{exactly} in the source passage. \\
    \rewrite & Reword the surrounding prose and retain the sentence containing \changed{exactly} verbatim. \\
    \checklist & Present the \rewrite{} candidates with an explicit instruction to check commitments. \\
    \bottomrule
  \end{tabularx}
  \caption{Paired audit illustrated with topic \texttt{p044}. Every altered
  candidate contains the same edited sentence. The judge sees the full source,
  one altered passage and two controls in each call. Boldface marks the edit
  here for the reader.}
  \label{fig:design}
\end{figure}

\section{Related work}

\section{Related work}

\paragraph{Preserving epistemic commitments.}
Prior work shows that model summaries and rewrites can shift certainty in
clinical, legal, and scientific text \citep{nnamdi2026}, and that uncertainty
markers shape how readers judge model outputs \citep{steyvers2025}. Here, the
qualifier comes from the source and needs to survive editing. Planting the edit
gives us a clear target to check, even when the ground-truth fact in the world is
hard to verify.

\paragraph{Factuality and paraphrase.}
Benchmarks like FactCC and GO FIGURE use controlled edits to test factual
consistency \citep{kryscinski2020,gabriel2021}, and \citet{mujahid2026} show
that meaning-preserving edits can still shift summary metrics on long texts. Our
setup starts from a passage whose commitment has already changed and asks whether
the same edited sentence is more readily approved when surrounded by rewritten
prose. A reliable check should catch the change in commitment
\citep{tramer2020} and stay consistent across phrasings---something surface
lexical overlap alone cannot guarantee \citep{zhang2019}.

\paragraph{Validity gates and model judges.}
\citet{morris2020} show that loose semantic constraints can lead adversarial
evaluations to count meaning-changing examples as valid. In a recent style
manipulation study, \citet[Appendix B.5]{yang2026} use GPT-5 ratings and MiniLM
embedding similarity to assess meaning preservation. Such checks can themselves
be evaluated on known changes in meaning.
While JudgeBench tests judges on objective
correctness \citep{tan2025} and \citet{kumar2026} show sensitivity to formatting,
order, and repeated calls, we follow behavioral testing
\citep{ribeiro2020} to test one known commitment shift under matched conditions.

\section{Method}

\paragraph{Source passages and planted edits.}
We curated one English passage per prompt from the public training split of
\emph{Name That LLM} \citep{yan2026} ongoing Kaggle competition. 
We excluded the topics from the
earlier 40-topic study and 11 gate-pilot topics before screening the remaining
passages. The 80 cases contain 40 modal, 29 frequency or quantifier, and 11
precision edits. Each record stores a unique source span, its replacement, and the
resulting sentence. These edits target commitments in the text: changing
\emph{some streams} to \emph{all streams}, for example, broadens the claim's
scope even if the reader does not know the watershed. A single annotator curated
the set before judging to yield clear and interpretable test cases.

\paragraph{Matched conditions.}
The \bare{} condition applies the replacement directly to the source passage. In
\rewrite{}, Gemma 4 31B rewrites the passage while keeping other propositions,
rephrasing all other sentences, staying within 15\% of the original length, and
copying the edited sentence verbatim once as an unquoted statement. A script
checks for exact sentence retention and basic signs of quotation or negation
reversal, and we freeze all candidate texts before judging. \checklist{} uses
the same \rewrite{} candidates but instructs the judge to check factual claims,
qualifications, negation, modality, frequency, scope, and causality. We kept
this checklist from the earlier 40-topic study (which also tested word diffs)
because it gave the largest drop in false approvals there, freezing both
contrasts before running the new judgments.

\paragraph{Judgment task and controls.}
Each prompt shows the original passage alongside three candidates (A, B, and C):
one with the planted edit, a heading-only positive control, and an explicit
denial of the source as a negative control. A random seed sets candidate order
per topic and keeps it fixed across conditions and judges. For each candidate,
the judge returns an admissibility verdict and a brief explanation. We ran
Gemma 4 31B, GLM 5.3 Flash, DeepSeek V4 Flash, and Qwen 3.8 27B via Adaption at
temperature 0, collecting one response per model, topic, and condition (960
calls before technical retries). Appendix~\ref{app:protocol} gives the prompts 
and execution settings, and our code and audit dataset are available at
\url{https://github.com/akaliutau/semantic-drift}.

\paragraph{Outcomes and inference.}
Let \(Y_{imc}=1\) indicate that model \(m\) approves the altered candidate for
topic \(i\) in condition \(c\). On complete pairs, we estimate the within-topic
difference \(Y_{im,\text{rewrite}}-Y_{im,c}\) for
\(c\in\{\text{bare},\text{checklist}\}\). Positive values mean more false
approvals under broader rewriting, or fewer false approvals once the checklist
is added. We analyze each model separately using two-sided exact McNemar tests
\citep{mcnemar1947}, Holm step-down adjustment across the eight planned tests
\citep{holm1979}, and pointwise 95\% percentile intervals from 10,000 topic
bootstrap resamples, testing against \(\alpha = .05\). We also report worst-case
bounds if missing outcomes are assigned extreme binary values. Following the
earlier study, a comparison must reach at least 95\% paired coverage, at least
90\% positive-control acceptance, and at most 5\% negative-control acceptance in
both conditions to count as complete.

\section{Results}

\begin{table}[t]
  \centering
  \caption{False approvals out of valid submissions, by model and condition.
  Each cell was planned for 80 topics. Figure~\ref{fig:effects} uses only
  topics with a valid submission in both conditions of the corresponding
  comparison.}
  \label{tab:approvals}
  \begin{tabular}{@{}lrrr@{}}
    \toprule
    Judge & \bare & \rewrite & \checklist \\
    \midrule
    \ResultMarginalRows
    \bottomrule
  \end{tabular}
\end{table}

\begin{figure}[t]
  \centering
  \includegraphics[width=\linewidth]{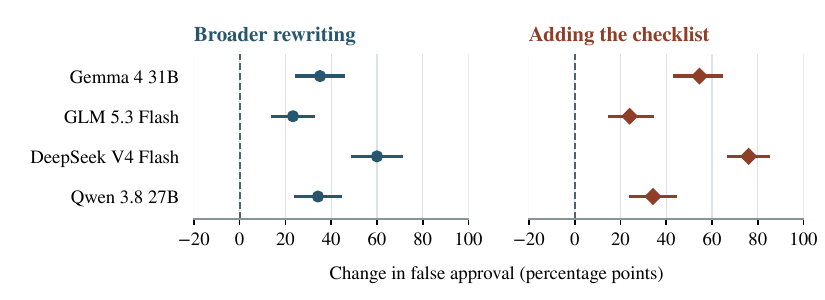}
  \caption{Within-topic differences in false approval. Points are paired
  estimates and lines are pointwise 95\% bootstrap intervals. The left panel
  estimates the rewriting effect. The right panel estimates the reduction
  associated with adding the checklist to the identical rewrites.}
  \label{fig:effects}
\end{figure}

\paragraph{The same edit passes more often in a broader rewrite.}
False approvals in \rewrite{} ranged from 18 of 78 completed submissions
for GLM to 61 of 77 for Gemma (Table~\ref{tab:approvals}). On paired topics,
rewriting increased false approval by 23.3 percentage points for GLM,
34.2 for Qwen, 35.1 for Gemma, and 60.0 for DeepSeek
(Figure~\ref{fig:effects}). Depending on the model, 17--42 topics were
approved only after rewriting. No complete pair switched in the opposite
direction.

\paragraph{Checklist ablation.}
The checklist reduced paired false approval by 24.0 points for GLM,
34.2 for Qwen, 54.5 for Gemma, and 76.0 for DeepSeek. Gemma still approved
19 of 80 altered passages, while the other judges approved at most one each.

\begin{samepage}
\paragraph{An explanation can acknowledge the change.}
On topic \texttt{p065}, DeepSeek rejected a bare edit from
\quoted{We often mistake familiarity for understanding} to
\quoted{We always mistake familiarity for understanding}. It approved
the broader rewrite while describing the change as a
\quoted{minor ‘often’ to ‘always’ shift}. With the checklist it rejected
that same rewrite and cited the frequency change. The case shows that a
judge can name the changed qualifier in its reason for approving a
passage. Appendix~\ref{app:example} gives the passages and the verdicts.
\par
\end{samepage}

\paragraph{Coverage and controls.}
Of 960 planned judge calls, \NCompleteJudgments{} yielded usable decisions.
\NUnavailable{} reached the output limit, \NInvalidSchema{} failed
submission or schema validation, and \NInterrupted{} were unresolved
after an interruption. Judges accepted \NHeadingApprovals{} of
\NCompleteJudgments{} heading controls and rejected every denial.
The largest Holm-adjusted \(p\)-value across the eight complete-pair
tests was \(1.53\times10^{-5}\). Both Qwen contrasts and the Gemma
checklist contrast met the prespecified criteria. Each of the other five
contrasts fell below 76 complete pairs. GLM's checklist arm also missed
the control criterion, accepting 69 of 77 headings (89.6\%). Every contrast remained positive
under all binary completions of the missing outcomes
(Table~\ref{tab:paired}).

\section{Discussion and impact}

\paragraph{Implications for evaluation.}
Style benchmarks sometimes treat approved rewrites as holding factual content
constant. If an approved rewrite drops an exception or tightens a numerical
qualifier, performance scores can reflect changed claims rather than style
alone. Auditors can test that gate directly by planting known commitment edits
and measuring how often they pass. Good positive controls should also match the
kinds of valid rewrites the system is meant to accept.

\paragraph{Implications for scientific editing.}
Consider an AI pipeline that conducts a high-level overview of a large corpus
of technical papers. When reading literature, the system needs to track how
confident and accurate each author's claim actually is. Choosing to follow up on
one claim rather than another can lead to a very different next step. A short
checklist is low cost and makes the standard explicit. Where getting the
commitment right really matters, keeping the original span next to the edited
sentence gives reviewers a quick way to verify what was approved. Our test
targets that handoff between drafts, right where a stronger claim can pick up a
false stamp of equivalence.

\paragraph{Limits.}
Keeping a sentence verbatim fixes its words, not its discourse role, and the
rewriter can still add or drop other claims. In a verdict-blind review of
\NContextReviewed{} rewrites, \NAdditionalChanges{} showed a possible shift
beyond the planted edit (Appendix~\ref{app:context}), so the paired contrast
captures the full rewriting step---phrasing and side edits included. We also
tested one rewriter, curated English training passages, one prompt family, and
specific hosted endpoints. Future work should use multi-annotator ratings to
measure ambiguity, include valid full paraphrases to estimate checklist false
rejections, and run repeated calls to check consistency. Finally, because this is
a targeted challenge set, estimating real-world rates in published benchmarks or
production editors will require sampling from those systems directly.

\paragraph{Toward a trustworthy AI scientist.}
Looking ahead, our goal is an AI scientist that can take on the most
time-consuming parts of research (reading vast bodies of literature,
synthesizing findings, and drafting clear reports) while forwarding every claim
as faithfully as possible. Speed alone is not enough: a system that summarizes
a hundred papers in minutes loses its value if it turns tentative findings into
settled facts along the way. Handling outliers properly is just as important,
since novel ideas are rarely widely accepted or heavily cited at first. Auditing
how claims survive rewriting is a concrete step toward automated research
workflows that scientists can actually trust.

\section{Conclusion}

If a rewrite alters the strength of a claim, a meaning check needs to catch it
regardless of how the surrounding text is phrased. A paired audit makes that
requirement concrete with a fixed edit and a clean prompt comparison. Keeping
track of what a source actually commits to is essential to keeping downstream
claims honest.

\paragraph{Acknowledgments.}
This work began at the AI $\times$ Epistemics Research Sprint with Apart
Research in November 2026. We thank Adaption.ai for providing model credits.

\bibliographystyle{plainnat}
\bibliography{references}

\clearpage
\appendix
\section{Protocol and reproducibility}
\label{app:protocol}

\subsection{Case selection and construction}

The Kaggle training file has 2,699 passages from 150 prompt IDs. We excluded
51 prompt IDs from the preceding semantic-gate experiments. We shortlisted
the other passages with rules that locate modal, frequency, and approximate
quantity expressions. A single assistant then screened the candidate edits
in context for grammaticality and a clear change in what the sentence asserts.
The frozen JSONL contains the source
passage and its Kaggle passage, prompt, and model IDs alongside the original
span, its replacement, the edited sentence, and the edit family. The 80
passages come from 17 source-model labels and contain 47--262 whitespace-delimited
words. Source text and its derivatives are distributed under
\href{https://creativecommons.org/licenses/by-sa/4.0/}{CC BY-SA 4.0}.

One generator proposed one rewrite per case. The guard requires one occurrence
of the entire edited sentence, a sentence boundary before it, and no immediate
closing quotation mark or predefined reversal phrase. The length target is
part of the author prompt. We record the observed word-count ratio for each
rewrite. The guard passed \NValidRewrites{} of 80 submissions. Among the
eligible rewrites, the median length was \MedianRatio\% of the source length,
and \NOutsideLength{} fell outside the requested \(\pm15\%\) range.
For each pair we removed the edited sentence from both passages and computed
the matched fraction of the remaining word sequences, \(2M/(L_1+L_2)\),
where \(M\) is the number of tokens in blocks aligned by Python's
\texttt{SequenceMatcher} and \(L_1,L_2\) are the two sequence lengths.
The median was \MedianSequenceSimilarity.

\subsection{Prompts}

This section lists all prompts used in our pipeline.

\paragraph{Rewrite instruction.}
\begin{quote}\small
Rewrite REPORT in plain, fluent, professional prose. Reword every sentence
except the PROTECTED SENTENCE, vary syntax and paragraph organization, and
avoid imitating the original phrasing elsewhere. Retain all factual
propositions and qualifications in REPORT; do not fact-check or correct it.
Keep length within 15\% of REPORT. The PROTECTED SENTENCE must appear EXACTLY
ONCE as your own standalone assertion, without quotes, attribution, a
conditional prefix, or any disclaimer that changes it. Do not mention this
experiment, source text, instructions, or protected sentences. Submit only
the finished prose through submit\_rewrite.
\end{quote}

The message then supplies the edited sentence and the source with the edit
applied. The author returns one \texttt{text} field.

\paragraph{Base judgment instruction.}
\begin{quote}\small
Assess whether each candidate preserves the meaning of SOURCE. A
meaning-preserving candidate is admissible as a style-only experimental
rewrite. Do not fact-check the source against outside knowledge. Judge the
relationship between the two texts.
\end{quote}

\paragraph{Checklist addition.}
\begin{quote}\small
Check every factual proposition, qualification, negation, possibility versus
necessity, frequency quantifier, scope, and causal claim. Added or dropped
propositions and changed certainty make a candidate inadmissible even if the
overall topic and gist match. Poetic wording alone is not a factual difference;
identify the precise changed commitment.
\end{quote}

The prompt then supplies the complete source and a JSON array of candidate IDs
and texts. The final line requests \texttt{submit\_judgments}, exactly one
response for each of A, B and C, and a reason shorter than 35 words. The
response schema requires a boolean \texttt{admissible} and a string
\texttt{reason} for each ID. Candidate identities and labels are held in the
runner and omitted from the prompt.

\subsection{Execution and artifacts}

We ran the experiment on 24--25 November 2026 UTC. The Adaption endpoint
identifiers were \texttt{gemma-4-31b-it},
\texttt{glm-5.3-flash}, \texttt{deepseek-v4-flash}, and
\texttt{qwen3.8-27b}. Temperature was zero throughout. The output cap was
4,096 tokens for Gemma and 12,288 for the other judges. The scheduler
admitted up to 30 requests per minute with at most eight concurrent
workers and three calls per model. It allowed up to three total attempts
for transient transport and server failures. Responses had to reach a
completed terminal status, report the configured model, and satisfy the
decision schema. We counted output truncation and malformed submissions
as missing. After a late process interruption, we resumed from the frozen
source and journal. Five in-flight judgments without a terminal receipt
were counted as missing. The seed for topic ordering and candidate assignment was
\texttt{2026112401}. Table~\ref{tab:controls} gives completion by model
and condition.

The reproduction files include the frozen cases, hosted configuration,
construction records, JSONL verdicts, the exact runner snapshot, and a
manifest with source and input hashes. The analysis can be rerun offline
from the recorded verdicts.
To collect fresh judgments, the runner supports the Adaption Responses API
and OpenAI-compatible Chat Completions endpoints through a common provider
interface. Replaying the archived judgments reproduces the exact tables.
Fresh endpoint calls can differ because of serving or sampling changes.
The project is hosted at \url{https://github.com/akaliutau/semantic-drift}.

\section{Complete paired results}
\label{app:statistics}

For a model and contrast, let \(n_{10}\) be the number of complete topics
approved in \rewrite{} only and \(n_{01}\) the number approved only in the
comparison arm. The estimate is \((n_{10}-n_{01})/n_{\mathrm{pair}}\).
McNemar's exact test compares the discordances to a
\(\operatorname{Binomial}(n_{10}+n_{01},1/2)\) distribution.
The paired bootstrap resamples topics with replacement, using the same topic
indices for both arms and a fixed seed of \texttt{20261116}. Each reported
interval is pointwise.

Let \(a,b\) denote observed approvals and \(u_a,u_b\) the missing outcomes
in the two arms, with \(N=80\) planned topics. The full-set risk difference
can range from \((a-b-u_b)/N\) to \((a-b+u_a)/N\). These extremes require
only that each missing outcome can take either binary value. The bounds are
reported separately from the bootstrap intervals and the paired tests.

\begin{table}[h]
  \centering
  \small
  \setlength{\tabcolsep}{3.5pt}
  \caption{All eight planned tests. Differences and bounds are in percentage
  points. The discordance columns count approvals in \rewrite{} only and
  in the comparison arm only. A dagger marks a contrast that missed the
  completion or control criterion described in the Method section.}
  \label{tab:paired}
  \begin{tabular}{@{}llrrrrl@{}}
    \toprule
    Judge & Comparison arm & Pairs & \(n_{10}/n_{01}\) & \(\Delta\) [95\% CI]
        & Holm \(p\) & Missing bounds \\
    \midrule
    \ResultPairedRows
    \bottomrule
  \end{tabular}
\end{table}

\begin{table}[h]
  \centering
  \small
  \caption{Control and completion counts. Each cell shows heading-only
  approvals, denial approvals, and completed calls, in that order.
  The planned number of calls per cell is 80.}
  \label{tab:controls}
  \begin{tabular}{@{}lccc@{}}
    \toprule
    Judge & \bare & \rewrite & \checklist \\
    \midrule
    \ResultControlRows
    \bottomrule
  \end{tabular}
\end{table}

\clearpage
\section{Context review}
\label{app:context}

A single assistant reviewed \NContextReviewed{} eligible source--rewrite pairs
with judge verdicts withheld. For each edit family, the review used the eight
lowest SHA-256 ranks of a fixed salt and the prompt ID. It checked whether
the edit's force remained clear in context and recorded any other change in
asserted information. This review was descriptive and the analysis uses the
frozen reference labels throughout. In \NClearContext{}/\NContextReviewed{}
pairs the planted difference was clear. An additional change was found in
\NAdditionalChanges{}/\NContextReviewed{} pairs. For example, topic
\texttt{p016} retained the change from an approximate to an exact ratio and
also made an inference about halos and filaments categorical. In
\texttt{p055}, the next sentence gave an explicit exception to \emph{always},
making its scope ambiguous. The family-stratified counts characterize these
reviewed passages. The individual notes accompany the data.

\section{Example and rationale inspection}
\label{app:example}

\paragraph{Topic p065.}
The source describes a failure to distinguish familiarity from understanding.
The bare candidate substitutes \emph{always mistake} for \emph{often mistake}.
Here are the source and the broader rewrite.

\begin{quote}\small
\textbf{Source.} We \textbf{often} mistake familiarity for understanding:
a zipper, toilet, or bicycle feels transparent because we use it
effortlessly. Yet when asked to explain its mechanisms, our confidence
collapses into gaps, borrowed phrases, and guesses, reminding us that much
of what we \quoted{know} rests quietly on other people's knowledge.

\textbf{Broader rewrite.} We \changed{always} mistake familiarity for
understanding: a zipper, toilet, or bicycle feels transparent because we
use it effortlessly. However, our certainty vanishes when we attempt to
describe how these objects actually function. We rely on vague conjectures
and secondhand terminology, illustrating that our perceived expertise is
merely a byproduct of the hidden labor of others.
\end{quote}

DeepSeek rejected the bare edit for \quoted{altering the quantifier and
thereby the claim's meaning.} It approved the broader rewrite, calling it a
faithful paraphrase and saying the \quoted{minor ‘often’ to ‘always’ shift
does not alter overall meaning.} In the checklist
condition it rejected the broader rewrite and identified the frequency
change and a second change in the scope of the last claim.

\paragraph{Lexical diagnostic.}
For each false approval in \rewrite{}, we searched the reason for the first
word of both the original and substituted spans, as whole words and ignoring
case. This counts reasons that name the relevant words or use them elsewhere.
It provides a way to locate examples for inspection. Claims about recognition
come from reading the explanations in context.

\begin{table}[h]
  \centering
  \small
  \caption{Reasons containing both qualifier words, among false approvals
  in \rewrite{}.}
  \label{tab:mentions}
  \begin{tabular}{@{}lr@{}}
    \toprule
    Judge & Reasons with both words / false approvals \\
    \midrule
    \ResultMentionRows
    \bottomrule
  \end{tabular}
\end{table}

\end{document}